\documentclass{article}

\usepackage[preprint]{nips_2018}

\usepackage{algpseudocode,algorithm,algorithmicx}  

\usepackage{amsfonts}  % blackboard math symbols
\usepackage{amsmath}
\usepackage{amssymb}
\usepackage{amsthm}
\usepackage{appendix}
\usepackage{bbold}
\usepackage{booktabs}  % professional-quality tables
\usepackage{color}
\usepackage{courier}
\usepackage{enumitem}
\usepackage{environ}
\usepackage{etoolbox}
\usepackage{float}
\usepackage{graphicx}
\usepackage{helvet}
\usepackage{mathtools}
\usepackage{mdwlist}  % single-spaced itemize* and enumerate*
\usepackage{microtype}  % microtypography
\usepackage{multirow}
\usepackage{natbib}
\usepackage{nicefrac}
\usepackage{nicefrac}  % compact symbols for 1/2, etc.
\usepackage{subfigure}
\usepackage[T1]{fontenc}  % use 8-bit T1 fonts
\usepackage{tabularx}
\usepackage{times}
\usepackage{url}  % simple URL typesetting
\usepackage[utf8]{inputenc}  % allow utf-8 input
\usepackage{verbatim}
\usepackage{wrapfig}
\usepackage{xspace}

\usepackage{hyperref}  % hyperlinks

\setlist{nosep}

\hypersetup{colorlinks = false, pdfborder = {0 0 0}}

\newif\ifdraft
\draftfalse

\renewcommand{\paragraph}[1]{\textbf{#1}\;\xspace}
\renewcommand{\eqref}[1]{Equation~\ref{eq:#1}}

\newcommand{\appref}[1]{Appendix~\ref{app:#1}}

\makeatletter
\newtheorem*{rep@theorem}{\rep@title}
\newcommand{\newreptheorem}[2]{\newenvironment{rep#1}[1]{\def\rep@title{#2 \ref{##1}}\begin{rep@theorem}}{\end{rep@theorem}}}
\makeatother

\newtheorem{theorem}{Theorem}
\newtheorem{lemma}{Lemma}
\newtheorem{corollary}{Corollary}

\newreptheorem{theorem}{Theorem}
\newreptheorem{lemma}{Lemma}
\newreptheorem{corollary}{Corollary}

\newif\ifreptheorem
\reptheoremfalse
\newif\ifshowproofs
\showproofsfalse
\NewEnviron{thm}[1]{
  \makeatletter
  \ifcsname defined:thm:#1\endcsname
    \reptheoremtrue
    \begin{reptheorem}{thm:#1}\BODY\end{reptheorem}
    \reptheoremfalse
  \else
    \begin{theorem}\label{thm:#1}\BODY\end{theorem}
    \csgdef{defined:thm:#1}{}
  \fi
  \makeatother
}
\NewEnviron{lem}[1]{
  \makeatletter
  \ifcsname defined:lem:#1\endcsname
    \reptheoremtrue
    \begin{replemma}{lem:#1}\BODY\end{replemma}
    \reptheoremfalse
  \else
    \begin{lemma}\label{lem:#1}\BODY\end{lemma}
    \csgdef{defined:lem:#1}{}
  \fi
  \makeatother
}
\NewEnviron{cor}[1]{
  \makeatletter
  \ifcsname defined:cor:#1\endcsname
    \reptheoremtrue
    \begin{repcorollary}{cor:#1}\BODY\end{repcorollary}
    \reptheoremfalse
  \else
    \begin{corollary}\label{cor:#1}\BODY\end{corollary}
    \csgdef{defined:cor:#1}{}
  \fi
  \makeatother
}
\NewEnviron{prf}[1]{
  \ifshowproofs
    \begin{proof}\BODY\end{proof}\label{app:prf:#1}
  \else
    \begin{proof}In \appref{prf:#1}.\end{proof}
  \fi
}

\newenvironment{titled-paragraph}[1]{\textbf{#1:}}{}

\newcounter{lineno}

\drafttrue

\title{Quit When You Can: Efficient Evaluation of Ensembles with Ordering Optimization}
\author{
  Serena Wang \\
  Google, Inc. \\
  \texttt{serenawang@google.com} \\
  \And
  Maya Gupta \\
  Google, Inc. \\
  \texttt{mayagupta@google.com} \\
  \And
  Seungil You \\
  Kakao Mobility \\
  \texttt{sean.you@kakaomobility.com} \\
}

\begin{document}

\maketitle

\label{document:begin}

\begin{abstract}
Given a classifier ensemble and a set of examples to be classified, many examples may be confidently and accurately classified after only a subset of the base models in the ensemble are evaluated. This can reduce both mean latency and CPU while maintaining the high accuracy of the original ensemble. To achieve such gains, we propose jointly optimizing a fixed evaluation order of the base models and early-stopping thresholds. Our proposed objective is a combinatorial optimization problem, but we provide a greedy algorithm that achieves a 4-approximation of the optimal solution for certain cases. For those cases, this is also the best achievable polynomial time approximation bound unless $P = NP$. Experiments on benchmark and real-world problems show that the proposed \emph{Quit When You Can} (QWYC) algorithm can speed-up average evaluation time by $2$x--$4$x, and is around $1.5$x faster than prior work. QWYC's joint optimization of ordering and thresholds also performed better in experiments than various fixed orderings, including gradient boosted trees' ordering.
%Speeding up a large ensemble using QWYC also achieved higher accuracy than training a smaller ensemble in the first place, which was not always true for prior work.
% SERENA KDD NOTES: the independently-trained vs. jointly-trained isn't that much of a focus of the experimental results anymore, so I'm removing it from the abstract.
%Comparisons of ensembles that have been independently-trained vs jointly-trained for the same problem show that it may be possible to speed up the evaluation of independently-trained ensembles around 2x more than jointly-trained ensembles, though jointly-trained ensembles may still be preferable for best accuracy.
\end{abstract}

\section{Introduction}
We consider the problem of efficiently evaluating a binary classifier that can be expressed as an ensemble of $T$  base models, $f(x) = \sum_{t= 1}^T f_t(x)$,
% \begin{equation}\label{eqn:ensemble}
% f(x) = \sum_{t= 1}^T f_t(x),
% \end{equation}
where each base model $f_t(x) \in \mathbb{R}$ contributes a score. 
The ensemble may have been trained:\\
\textbf{Independently:} Each $f_t$ is trained individually, for example, random forests \citep{Breiman:01}.\\
\textbf{Sequentially:} Each $f_t$ is trained taking into account fixed $f_1$, $f_2$ $\ldots,$ $f_{t-1}$, for example, boosted trees \citep{rfsRock:14}. \\
\textbf{Jointly:} All $f_t$ are trained together, for example, generalized additive models \citep{HastieTibshirani:90} or an ensemble of lattices \citep{canini:2016}.

Existing work on ensemble pruning can be generally categorized as static ensemble pruning or dynamic ensemble pruning. In traditional static ensemble pruning, the same pruned subensemble is used for all examples (\citep{Zhou:2002, MartinezMunoz:2006, Margineantu:1997, Tamon:2000}, and many others). In contrast, in dynamic ensemble pruning, a different subensemble may be used to evaluate each example \citep{Xiao:2010}.
% the idea being that different base models may be better suited for different subsets of examples. 

The approach that we propose is dynamic:
we propose to evaluate a larger subset of base models for examples that are more difficult to classify. The key idea is that if some examples are easy to classify, then such easy 
examples can be confidently and accurately classified after evaluating only a few base models, but for hard examples, we may choose to evaluate the full ensemble. This adapts the computation to the difficulty of the classification problem, reducing 
average latency and computational resources (e.g. CPU) needed to classify. This dynamic approach is complementary to the traditional static ensemble pruning approach, as the dynamic approach can act on a pre-pruned ensemble to achieve even further average speedups on a per-example basis.

A variety of dynamic ensemble pruning strategies exist that rely on different heuristics for selecting subensembles. \citet{Woods:1997} introduced the idea of selecting base models for different clusters of examples, and ordering the base models by accuracy within each cluster. \citet{Santana:2006} showed that ordering base models by both accuracy and diversity can actually produce more accurate subensembles at test time. \citet{Santos:2007} proposed a slightly different heuristic of choosing the most confident base models for each example. 
\citet{Fan:2002} suggested ordering by total benefits of the base models, where total benefits may include some combination of accuracy and the classifier's objective. The ultimate goal of these selection criteria is to achieve faster evaluation time without sacrificing too much performance compared to the full ensemble.

While ordering by total benefits, diversity, and confidence have been shown to work better than accuracy alone for achieving faster evaluation time, we aim to bypass these heuristics entirely by jointly optimizing the order with the pruning mechanism to directly improve the evaluation time objective without differing too much from the original ensemble. 

Specifically, to dynamically prune the ensemble, we propose an objective that defines an optimal ordering $\pi$ of the $T$ base models and an optimal set of $2T$ corresponding early decision thresholds 
$\{\epsilon^+, \epsilon^-\}$ that minimize the average number of base models evaluated, while maintaining
the accuracy of the full ensemble. We also propose a simple early stopping mechanism for a
given example $x$ as follows: after evaluating the $r$th base model, if the accumulated
sum $\sum_{t=1}^r f_t(x)$ is above the $r$th early positive threshold
$\epsilon^+_r$, then the example is classified as the positive class, and the
evaluation is stopped early (and vice-versa for negative examples). Otherwise, the $r+1$th base model is evaluated. 
%The proposed objective includes a hyperparameter to control how faithful the fast classifier should be to the full classifier, for representative examples. 

While it is computationally challenging to find a globally optimal solution for this combinatorial optimization problem, we propose a greedy algorithm which we refer to as \emph{Quit When You Can} (QWYC). For certain ensembles, QWYC achieves a 4-approximation of the optimal solution, which is also the best approximation bound that can be achieved in polynomial time unless $P = NP$. While we present QWYC with simple early stopping criteria using thresholds, other pruning mechanisms may be substituted into the QWYC algorithm to be optimized with the ordering.  

Experiments on both real-world and benchmark problems show that QWYC can speed up computation time by $2$x--$4$x while classifying faithfully
compared to the full ensemble on held-out examples. 
%Speeding up a large ensemble using QWYC also achieved higher accuracy in experiments than training a smaller ensemble in the first place, which was not always true for the most related prior solution (Sections \ref{sec:exp1} and \ref{sec:exp2}). For example, experiments on the UCI Adult dataset showed that using QWYC to reduce an ensemble of size 500 down to an average of 40 base models evaluated per example achieved higher accuracy than training an ensemble of size 40 to begin with (Figure \ref{fig:adult_accuracy}).
QWYC's joint optimization of ordering and early stopping criteria also performed better in experiments than various fixed heuristic orderings, including the natural ordering produced during training of gradient boosted trees (Section \ref{sec:experiments}).

Our main contribution is to demonstrate both the value and practicality of jointly optimizing ordering with early stopping criteria for dynamic ensemble pruning. 

\section{Related Work}\label{sec:related}

The most related prior work is the dynamic ensemble pruning strategy of \citet{Fan:2002}. The authors order the base models based on their individual utility. For example, they order from the base model that has the highest accuracy when used by itself to the base model that has the lowest accuracy when used by itself. They propose an early stopping criteria where the threshold to declare an early positive or negative class depends on a probabilistic modeling of the running score. A more detailed explanation and implementation is given in Section \ref{sec:fan}. In contrast, we take a discriminative approach: we propose to jointly optimize the ordering with the early stopping criteria to minimize the expected evaluation cost, constrained to producing the same classification decisions as the full classifier within a given error margin.

One common approach in dynamic ensemble pruning involves using clustering to match examples to their most accurate subensembles. These methods require the user to specify an ordering criteria for each cluster \citep{Woods:1997}. Some work has been done on improving the ordering of base models for each cluster by selecting for both accuracy and diversity instead of just accuracy \citep{Santana:2006, Xiao:2010}. Our method is thus complementary to these clustering approaches -- for examples in each cluster,  QWYC can choose an ordering that directly reduces evaluation time rather than relying on the user to specify selection heuristics.

A special case of the efficient ensemble evaluation problem is when the base 
models are assumed to be independently-trained and randomly-ordered, and each 
base model only contributes a vote of yes or no for each class, rather than a 
score.  Then the incomplete votes can be modeled as 
draws from a multinomial distribution \citep{HernandezLobato:2009, Schwing:2011} 
and the class priors can be incorporated \citep{Soto:2016}. Other researchers 
have shown similar results using a Gaussian approximation \citep{Kolda:2011}.  
In those papers, only intermediate thresholds for stopping early are chosen; 
the base models are not re-ordered as part of the speed-up.

A related but different problem is trying to adaptively minimize the set of features needed to evaluate each $x$, either to reduce costs or time associated with obtaining those features \citep{Parrish:2013, Wang:2015, Nan:2016}. 
In this paper, we do not explicitly model feature computation at all.  We depend only on the cost of evaluating each base model and treat each base model evaluation cost as independent of the cost of evaluating other base models. One could model feature evaluation costs via the base model evaluation cost, but if two base models use the same expensive feature, then our model would not take into account that the second use of the feature was essentially free. 

A more general version of the problem than what we tackle here is that instead of using the same optimized ordering of the base models for all examples, one could adaptively change the choice of which base model to evaluate next \citep{Koller:2011, Weinberger:2013}. Such instance-adaptive orderings could potentially offer even greater gains if some of the base models are relatively expensive to evaluate. However, these orderings would require substantially greater system complexity in practice.

Others have focused on how to design and train a cascade of base models to 
enable significant early rejections \citep{ViolaJones:2001, FloatBoost:2002,
Vasconcelos:2010}.  For example, \citet{ViolaJones:2001} produced a seminal real-time 
face recognition systems by hand-designing a simple-to-complex cascade of 
classifiers that could perform early rejection. In this
paper, we take the ensemble as given, and propose a method for speeding up a 
given already-trained ensemble.

Lastly, another strategy for fast evaluation of ensembles is to evaluate the 
base models in parallel. However, parallelization does not reduce the mean computational resources 
needed to classify, which is one of the goals of this work. Moreover,  parallelized evaluation is a complementary strategy: one can  parallelize sub-ensembles  of
base models such that each $f_t$ represents one sub-ensemble. In practice, if one can evaluate batches of samples at a time, we have found it is less complex in terms of software engineering and more efficient to simply parallelize over samples, rather than parallelizing over base models. 

\section{Optimization Problem}
We take as given a linearly-separable model $f(x): x \in \mathbb{R}^D \rightarrow \mathbb{R}$ where $f(x) = \sum_{t= 1}^T f_t(x)$,
an expected time (or other cost) $c_t \in
\mathbb{R^+}$ to evaluate the $t$th base model $f_t(x)$ for $t=1, \ldots, T$, and a decision threshold  $\beta \in \mathbb{R}$ for classifying $f(x)$. 

We propose optimizing a fast classifier over the set of all permutations of  $\Pi$ that map $\{1, 2, \ldots, T\}  \rightarrow \{1, 2, \ldots, T\}$, and positive and negative threshold vectors $\epsilon^+, \epsilon^- \in \mathbb{R}^T$  to minimize the expected evaluation cost for a random sample $X \in \mathbb{R}^D$ drawn from a data distribution $\mathbb{P}_X$ subject to $\epsilon^- \leq_+ \epsilon^+$ (here, $\leq_+$ is the element-wise inequality), with a constraint on what percent of samples can be classified differently by the fast classifier from the full classifier  $f(x)$ controlled by a misclassification rate  $\alpha$:
\begin{equation}
\begin{aligned}
& \underset{{\pi \in \Pi, \epsilon^- \in \mathbb{R}^T, \epsilon^+ \in \mathbb{R}^T}} {\text{minimize}}
& \mathbb{E}_X\left[  \sum_{t = 1}^{T}  c_{\pi(t)} I_{X, t, \pi, \epsilon^+, \epsilon^-}\right] \text{   subject to      } & \mathbb{E}_X\left[Z_{X, t, \pi, \epsilon^+, \epsilon^-}\right] \leq \alpha, \\
&&& \epsilon^- \leq_+ \epsilon^+
\end{aligned}
\label{eqn:Xobjective}
\end{equation}
where $I_{X, t, \pi, \epsilon^+, \epsilon^-}$ is a Bernoulli random variable that is $1$ if $X$ is not classified yet after $t$ base models in the ordering denoted by permutation $\pi$ according to the decision threshold vectors $\epsilon^+$ and $\epsilon^-$, and is $0$ otherwise; and $Z_{X, \pi, \epsilon^+, \epsilon^-}$  is a Bernoulli random variable that is $1$ if the fast classification of $X$ differs from the full evaluation classifier decision for $X$, and $0$ if the fast and complete classification decisions agree.

% MAYA-KDD NOTES: I found this figure actually kind of confusing because the arrows make it look like all the base models get run. And I think it's clear in the text. And we needed the space. So I'm commenting it out.
%\begin{figure}
%\includegraphics[width=\linewidth]{fast_eval_stage.png}
%\caption{Classification decision process for given ordering $\pi$ and thresholds at each stage, %$\epsilon^+, \epsilon^-$. The classification decision is made sequentially for a sample $x$. If %$f_{\pi(1)}(x)$ is greater than $\epsilon_1^+$ or less than $\epsilon_1^{-}$, then the positive or negative classification decision is made and we stop the evaluation (thus "quit"). Otherwise we proceed to the next stage, $t=2$, and repeat the procedure until we can quit the process or reach to $t=T$.}
%\end{figure}

\subsection{Empirical Approximation}
In practice, we do not have direct access to $\mathbb{P}_X$ and instead we take as given a set of $n$ examples $\{x_i\}$ assumed to be sampled independently and identically from $\mathbb{P}_X$. 
Let $g_r(x, \pi)$ denote the accumulated sum
(incomplete score) after $r$ base models ordered as per $\pi$: $
g_r(x, \pi) = \sum_{t = 1}^r f_{\pi(t)}(x)$.

After evaluating the $r$th base model, an example $x$ belongs to one of three
classes--positive, negative, or uncertain: $P_r := \{x | g_r(x, \pi) > \epsilon^{+}_r\}$, $N_r := \{x | g_r(x, \pi) <
\epsilon^{-}_r\}$, $U_r := \{x |
\epsilon^{-}_r \leq g_r(x, \pi) \leq \epsilon^{+}_r\}$ for $r \in \{1, \ldots, T \}$
and $U_0 = \mathbb{R}^D$. If $x \in P_r$ or $x \in N_r$, then $x$ is classified as a positive or negative class, respectively, and we terminate the evaluation at the $r$th step.
On the other hand, if $x \in U_r$, the classification decision is uncertain, and we continue to evaluate the $\pi(r+1)$th base model.
Define three more classes: $N_{\textrm{full}} := \{x | f(x) < \beta\}$,
$P_{\textrm{full}} := \{x | f(x) \geq \beta\}$, $C_r := \bigcap_{t=0}^{r}
U_t$ 
where $C_r$ is the set of samples that remain unclassified after evaluating the $r$th base model.

We approximate the objective (\ref{eqn:Xobjective}) with the empirical analogue:
\begin{equation}
\begin{aligned}
&\underset{{\pi \in \Pi, \epsilon^- \in \mathbb{R}^T, \epsilon^+ \in
\mathbb{R}^T}}{\text{minimize}}  && \frac{1}{N}\sum_{t = 1}^{T}c_{\pi(t)} \sum_{i=1}^N \mathbf{1}[x_i \in C_{t-1}]\\
&\text{subject to} & & \frac{1}{N} \sum_{i=1}^N  \mathbf{1}[x_i \in \bigcup_{t=1}^T \left\{C_{t-1}
\cap P_t \cap N_{\textrm{full}}\right\}] \\
&&&+ \frac{1}{N} \sum_{i=1}^N 
\mathbf{1}[x_i \in \bigcup_{t=1}^T \left\{C_{t-1} \cap N_t \cap P_{\textrm{full}}\right\}]  \leq \alpha \textrm{   and } \epsilon^- \leq_+ \epsilon^+.
\end{aligned}
\label{eqn:objective}
\end{equation}

In Section \ref{sec:optimize}, we give an algorithm to approximately optimize (\ref{eqn:objective}).
%\begin{align} \label{eqn:objective}
%& \arg \min_{\substack{\pi \in \Pi, \\ \epsilon^+ \in \mathbb{R}^T,\\ \epsilon^- \in \mathbb{R}^T}} \sum_{i=1}^N  \sum_{\tau = 1}^{T}  \bar{c}_{\pi(\tau)} I_{x_i, \tau, \pi, \epsilon^+, \epsilon^-}, \\ 
%\textrm{such that } & \\
%&\frac{1}{N} \sum_{i=1}^N FP_{x_i, \tau, \pi, \epsilon^+} \leq \alpha_{FP},\\
%&\frac{1}{N} \sum_{i=1}^N FN_{x_i, \tau, \pi, \epsilon^-} \leq \alpha_{FN},
%\end{align}
%where $I_{x, \tau, \pi, \epsilon^+, \epsilon^-}$ is the indicator for the event that $x$ required evaluating the $\pi(\tau)$ base model given the positive classification and negative classification thresholds $\epsilon^+$ and $\epsilon^-$:
%\begin{equation}
%I_{x, \tau, \pi, \epsilon^+, \epsilon^-} = \left(\displaystyle \bigcap_{r=1}^{\tau - 1}   \sum_{s = 1}^r f_{\pi(s)}(x)  \leq \epsilon_r^+ \right) \left( \displaystyle \bigcap_{r=1}^{\tau - 1} \sum_{s = 1}^r f_{\pi(s)}(x) \geq \epsilon_r^- \right),
%\end{equation}
%and $FP$ (and $FN$) are indicators for the event that the fast classifier decision for $x$ is a false positive (or false negative) compared to the complete classification $f(x)$:
%\begin{equation}\label{eqn:delta_FP}
%FP_{x, \tau, \pi, \epsilon^+}= \left( \displaystyle \bigcup_{r=1}^{T}  \sum_{s = 1}^r f_{\pi(s)}(x) > \epsilon_r^+   \cap  f(x) \leq 0 \right),
%\end{equation}
%\begin{equation}\label{eqn:delta_FN}
%FN_{x, \tau, \pi, \epsilon^-} = \left( \displaystyle \bigcup_{r=1}^{T}   \sum_{s = 1}^r f_{\pi(s)}(x) < \epsilon_r^-   \cap  f(x) > 0 \right).
%\end{equation}

\textbf{Representative Dataset:} Certainly the value of the solution to
(\ref{eqn:objective}) will depend on how representative the examples $\{x_i\}$
are of the examples that will be seen in actual usage of the model. Because
optimizing  (\ref{eqn:objective}) does not require labels for the examples, it
is often easier in practice to obtain a large number of unlabeled examples to use in
optimizing  (\ref{eqn:objective}).

\textbf{Filtering Candidates:}  An important use case in practice for fast classification is when the prior probability is for the negative class, and we wish to reject examples quickly. For example, when recommending products to a user from a database of millions of products, one hopes to quickly reject most products, but the most promising ones should be fully-scored, and then their classifier scores can be sorted to rank the remaining small set of top candidates. For such cases, the objective (\ref{eqn:objective}) can be simplified to only finding early-rejection thresholds $\{\epsilon_t^-\}$, as any positively classified samples will need the full score and must thus be fully evaluated.

% Additional use cases and extensions are given in
% \ref{sec:extensions}.

\section{QWYC: Optimizing the Order and Early-Stopping Thresholds}\label{sec:optimize} Since (\ref{eqn:objective}) is a combinatorial
optimization over $\pi$, coupled with $2T$-dimensional optimization over the real-valued threshold vectors $\epsilon^{-}$ and $\epsilon^{+}$, we have little hope of finding a global optimum except for small $T$. We propose a greedy optimization of (\ref{eqn:objective}) for both the ordering of the base models and the thresholds, which we refer to as \emph{Quit When You Can} (QWYC).

% Maya notes: you already made this point above.
%QWYC does not require having labeled examples, which makes it easier in practice to sample large sets of representative examples to use in optimizing (\ref{eqn:objective}).

The greedy optimization proceeds iteratively, 
first calculating the optimal thresholds to satisfy the constraints of (\ref{eqn:objective}) for each 
of the possible $T$ base models that could be selected as the first base model 
$\pi(1)$.  The  base model that minimizes the evaluation time ratio is then
set as $\pi(1)$ with its corresponding optimized thresholds. The process is then repeated for each of the base models in order, taking into account the previous chosen base models and thresholds.
The evaluation time ratio $J_t$ is defined as
$$J_t(\epsilon^+,\epsilon^-,\pi) = \frac{c_{\pi(t)} \sum_{i=1}^N \mathbf{1}[x_i \in C_{t-1}]}{n_{\pi(t)}}  $$
where  $n_{\pi(t)} = \sum_{i=1}^N \mathbf{1}[x_i \in C_{t-1} - C_t]$ is the number of additional examples for which submodel $\pi(t)$ is able to stop evaluating early. Intuitively, $J_t$ is the additional evaluation time required to stop evaluating early on each example.

Algorithm \ref{alg:1} details this procedure, where we have pulled out the QWYC
optimization of the thresholds into a subroutine detailed in Algorithm
\ref{alg:2}.
  Optimizing  (\ref{eqn:objective})  for
each $\epsilon^-_r$ can be done using binary search, since the objective is monotonically decreasing with respect to $\epsilon^-_r$, and the constraint is monotonically increasing with respect to $\epsilon^-_r$ (and analogous logic holds for each $\epsilon^+_r$). Assuming that the binary
search is $O(N)$ where $N$ is the size of the dataset (given a limited number of
iterations), the time complexity for this procedure is $O(T^2N)$. 
The QWYC early-stopping thresholds can be computed for any ordering of the base models, and in the experiments we compare  different ordering methods combined with QWYC early stopping thresholds.

\begin{algorithm}
\caption{Optimize Order}\label{alg:1}
\begin{algorithmic}[1]
% \State Let $J(\pi,\epsilon^+,\epsilon^-)$ be the value of the objective in (\ref{eqn:objective})
%\State Initialize $\epsilon^+_1,\ldots,\epsilon^+_T \gets \infty$,$\epsilon^-_1,\ldots,\epsilon^-_T \gets -\infty$
\State Initialize $\pi(i) \gets i$ for $i = 1,\ldots,T$
%\State Let $C(\pi,\epsilon^+,\epsilon^-)$ denote the value of the constraint in (\ref{eqn:objective}).
% \State Initialize ${\epsilon_i^-} \gets -\infty$, ${\epsilon_i^+} \gets \infty$ for $i = 1,\ldots,T$
\For {$r = 1$ to $T$}
\State Initialize $k^* \gets r$ 
\State Initialize $J_r^* \gets \infty$
\State ${\epsilon^+_r}^* \gets \infty,{\epsilon^-_r}^* \gets-\infty$
\For {$k = r$ to $T$}
\State Swap $\pi(r)$ with $\pi(k)$
\State ${\epsilon^+_r}',{\epsilon^-_r}' \gets$ Optimize
Thresholds($\pi$,$\epsilon^+$,$\epsilon^-$)
\If{$J_r(\pi,\epsilon^+,\epsilon^-) < J_r^*$}
\State $J_r^* \gets J_r(\pi,\epsilon^+,\epsilon^-)$
\State ${\epsilon_r^+}^* \gets {\epsilon_r^+}'$, ${\epsilon_r^-}^* \gets
{\epsilon_r^-}'$ \State $k^* \gets k$ 
\EndIf
\State Swap $\pi(r)$ with $\pi(k)$ \Comment{Revert $\pi$}
\EndFor
\State Swap $\pi(r)$ with $\pi(k^*)$, $ \epsilon_r^+ \gets  {\epsilon_r^+}^*$,
$\epsilon_r^- \gets  {\epsilon_r^-}^* $
\EndFor
\State \Return $\pi, \epsilon^+, \epsilon^-$
\end{algorithmic}
\end{algorithm}

% An outline of the OptimizeThresholds algorithm referenced in Algorithm \ref{alg:1} is given here. 

\begin{algorithm}
\caption{Optimize Thresholds}\label{alg:2}
\begin{algorithmic}[1]
\State $\textbf{Input}$: Index $r$ to optimize
\State $\textbf{Input}$: $\epsilon^-_1, \epsilon^-_2 \ldots, \epsilon^-_{r-1}$
and $\epsilon^+_1, \epsilon^+_2 \ldots, \epsilon^+_{r-1}$ \State $\textbf{Input}$: $\pi(1), \pi(2), \ldots, \pi(r)$
\State Optimize (\ref{eqn:objective}) with $T = r$ for $\epsilon^-_r$ using binary search.
\State Optimize (\ref{eqn:objective}) with $T = r$  for $\epsilon^+_r$ using binary search.
\State \Return $\epsilon^+_r,\epsilon^-_r$
\end{algorithmic}
\end{algorithm}

\subsection{QWYC Approximation Bound}
In certain cases, the problem of finding the optimal ordering and early stopping thresholds reduces to the Pipelined Set Cover problem \citep{Munagala:2005}, and the QWYC greedy algorithm therefore achieves a 4-approximation of the optimal evaluation time. 
% Intuitively, when we evaluate each base model $f_t$, each $f_t$ allows us to stop early on a set of examples $S_t$. We finish evaluating all examples in the training dataset when the sequence of submodels ${f_t}$ have covered all of the examples. In the pipelined set cover problem, each $f_t$ covers a set $S_t$, and the goal is to order the $f_t$'s to cover the full example set with minimum cost. 

Let $S_t(i)$ be the maximal set of examples for which base model $f_t$ can make an early negative or early positive classification while satisfying the constraint when $\alpha = 0$ and $\pi(i) = t$ after a binary search over $\epsilon^+_i, \epsilon^-_i$ (assuming no early stopping before reaching the $i$th base model). For $i \geq 1$, $S_t(i) \supseteq S_t(1)$. Depending on the base models that came before the $i$th base model, $S_t(i)$ may contain more elements than $S_t(1)$. Let $OPT$ be the optimal cost under the conservative restriction that $S_t(i) = S_t(1)$ for $i \geq 1$. Let $OPT^*$ be the true optimal cost. Define \textsc{Pipeline} to be the set of problems where $OPT  \leq OPT^* + C$, such that the optimal cost that restricts early stopping to $S_t(1)$ differs by a constant from the true optimal cost. While this constant may seem arbitrary, we provide a concrete example of a problem from \textsc{Pipeline} in Appendix \ref{sec:pipeline-ex}.

\begin{thm}{pipeline}
For problems in \textsc{Pipeline}, the QWYC algorithm achieves a 4-approximation of the optimal cost, which is also the best achievable approximation bound unless $P = NP$.
\end{thm}

\begin{prf}{pipeline}
% Show that QWYC achieves 4 approx of OPT, and therefore OPT*.
Let $\alpha = 0$ and restrict $S_t(i) = S_t(1)$ regardless of position $i$.
Let $OPT$ be the evaluation time cost of the optimal ordering and thresholds under these restrictions. Finding the optimal $\pi,\epsilon^+,\epsilon^-$ under these restrictions then maps directly onto the Pipelined Set Cover problem, where the goal is to find the optimal ordering $\pi$ such that the ordered sequence of sets $S_{\pi(1)}(1), ..., S_{\pi(T)}(1)$ covers all possible training examples contained in $\cup_{t=1}^T S_t(1)$ while achieving the lowest total evaluation time cost. 

% Let $U = \{e_1,...,e_N\}$ be the set of examples in the training dataset. Let $\{S_1,...,S_T\}$ be the sets of examples that are classified correctly by submodels $\{f_1,...,f_T\}$ respectively. Let the objective be the same as defined in equation \ref{eqn:Xobjective}. When $\alpha = 0$, this objective is equivalent to $\sum_{t=1}^T c_{\pi(t)} |U - \cup_{j=1}^{t-1} S_{\pi(j)}|$. 

Theorem 1 by \cite{Munagala:2005} shows that the Greedy algorithm that chooses the set that minimizes the cost ratio each time achieves a 4-approximation of $OPT$. Since QWYC minimizes the evaluation time ratio $J_r$ when choosing the $r$th base model and early stopping thresholds $\epsilon^+_r and \epsilon^-_r$, QWYC achieves a 4-approximation of $OPT$.

Since in $\textsc{Pipeline}$, $OPT \leq OPT^* + C$, QWYC also achieves a 4-approximation of $OPT^*$ for problems in $\textsc{Pipeline}$. 

\cite{Feige:2004} also show that problems in $\textsc{Pipeline}$ with uniform cost do not admit to better than a 4 approximation unless $P = NP$.
\end{prf}

\section{Experiments}\label{sec:experiments}
We demonstrate the performance of QWYC's joint optimization of ordering and early stopping thresholds on four datasets detailed in  Table \ref{tab:datasets} (in Appendix): two benchmark datasets, and two real-world case studies from a large internet services company. For datasets without a pre-defined train/test split, we randomly shuffle the dataset and split it 80-20 to produce a train and test set. The train set is used to optimize the ordering and early stopping thresholds.

Let QWYC* refer to Algorithm \ref{alg:1}, our proposed joint optimization of ordering and early stopping thresholds. To illustrate the effectiveness this joint optimization, we compare QWYC* to the baseline of fixing a variety of pre-selected orderings while optimizing the early stopping thresholds using Algorithm \ref{alg:2}. We describe these pre-selected orderings in Appendix \ref{sec:order_comparisons}.

We also compare QWYC* to the suggested ordering and early stopping mechanism proposed by \citet{Fan:2002}. \citet{Fan:2002} suggest pre-selecting an ordering by individual total benefits, and propose an early stopping mechanism that is more complex and flexible than Algorithm \ref{alg:2}. We describe their early stopping mechanism in detail in Appendix \ref{sec:fan}. They use Individual MSE as a simple total benefits metric. Let Fan* refer to the early stopping mechanism of \citet{Fan:2002} with pre-selected Individual MSE ordering. In addition to Individual MSE, for every pre-selected ordering we try with Algorithm \ref{alg:2}, we also try it with the \citet{Fan:2002} early stopping mechanism.

%For completeness of evaluating these orderings, we also compare to hybridized versions of the \citet{Fan:2002} algorithm that mix their early stopping thresholds with other pre-specified orderings. We describe in detail the orderings we compare to in Appendix \ref{sec:order_comparisons}.

Overall, we compare QWYC*'s joint optimization of ordering and thresholds to:

\begin{enumerate}
\item Fan*: the suggested ordering and early stopping method of \citet{Fan:2002}.

\item Combining different pre-selected orderings with our simple early stopping thresholds (Algorithm \ref{alg:2}) and the more flexible early stopping mechanism proposed by \citet{Fan:2002}.

\item Training a smaller ensemble, but always fully evaluating it.
\end{enumerate}

To illustrate the tradeoff between fast evaluation and faithfulness to the full ensemble, we vary the $\alpha$ hyperparameter that controls the maximum \% classification differences from the full ensemble for experiments that use either the QWYC* joint optimization or Algorithm \ref{alg:2} with other pre-selected orderings. Similarly, for comparisons that use the \citet{Fan:2002} early stopping mechanism, we vary the $\gamma$ hyperparameter that indirectly controls the \% classification differences from the full ensemble. 

\textbf{Benchmark Experiments Set-up:}
For the Adult and the Nomao benchmark datasets from the UCI Machine Learning Repository \citep{UCI}, we train gradient boosted tree (GBT) ensemble models \citep{friedman}. A GBT ensemble model is an additive model where the output is the sum of $T$ regression trees. The training process produces a natural ordering of the trees because each tree is added to the ensemble sequentially, which we compare to in experiments. When training the full ensemble, we perform hyperparameter optimization on a validation set over the number of base trees in the ensemble, the maximum depth of a single base tree, and the learning rate. We treat the evaluation cost of each base model as a constant $c_t = 1$ for all $t$, which is a conservative assumption as the maximum depth of each base tree is bounded.

\textbf{Experiment 1: UCI Adult Dataset:}
This is a binary classification task where the objective is to predict whether or not a person's income is greater than \$50,000. The full ensemble has $T=500$ base trees of maximum depth 5 and acts on a total of $D=14$ features. We use the predefined train/test split provided in the repository.

\textbf{Experiment 2: UCI Nomao Dataset:}
This is a binary classification task where the goal is dedpulication: each labeled example contains information about 2 sources, and the label denotes whether those 2 sources refer to the same business. The full ensemble has $T=500$ base trees of max depth 9 and uses the strongest $D=8$ features out of 120 available features. 

\begin{figure}
\includegraphics[width=\linewidth]{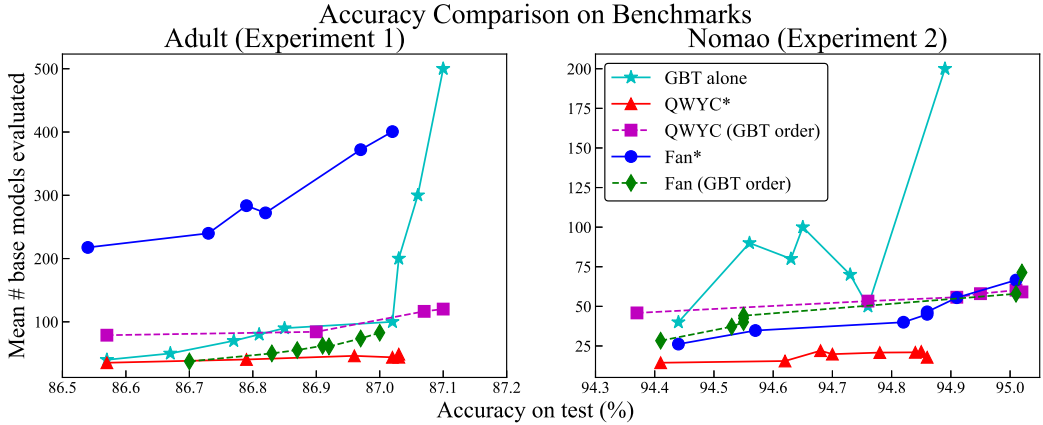}\caption{Test accuracy vs. mean \# base models evaluated for Adult and Nomao experiments. QWYC (GBT order) refers to Algorithm 2 with pre-selected GBT ordering, and Fan (GBT order) refers to \citet{Fan:2002} early stopping mechanism with pre-selected GBT ordering.}
\label{fig:nomao_adult_acc}
\end{figure}

We compare the performance of QWYC* to the \citet{Fan:2002} early stopping method with their suggested pre-selected ordering by Individual MSE. We also compare QWYC* to a baseline of training smaller gradient boosted tree ensembles. 
%When training the smaller gradient boosted tree ensembles, we keep the maximum depth of a single base tree the same as that of the original larger ensemble and perform hyperparameter optimization on a validation set over the learning rate. 
Finally, since gradient boosted trees produce a natural ordering of base models in the training process, we compare to the additional baseline of using the original gradient boosted tree ordering as a pre-selected ordering.
%for both the \citet{Fan:2002} early stopping method and the simple early stopping method in Algorithm \ref{alg:2}.
Figure \ref{fig:nomao_adult_acc} shows the tradeoff between accuracy and the number of base models evaluated for QWYC*, Fan*, and these baselines. 

\textbf{Real-World Experiments Set-up:}
The two real-world case studies from a large internet services company are both Filter-and-Score problems: the goal is to quickly filter the negative examples, and if the algorithm makes a positive classification then the full ensemble $f(x)$ is required for further processing by later stages. Thus, only $\epsilon^{-}$ is optimized. Note that the fact that the sizes of the train and test datasets are fairly small (see Table \ref{tab:datasets}) is not indicative of the need for fast evaluation. For example, real-world case study 1 requires evaluating the model over one trillion examples a day in response to real-time requests. 

For the real-world case studies, we were given ensemble models where the base models are lattices, which are interpolated look-up tables (LUTs), and can be trained with the open-source package TensorFlow Lattice \citep{canini:2016}. See Appendix \ref{sec:lattice} for more details on lattice ensembles. We were given jointly-trained ensembles, but we re-trained each ensemble with independent-training of the base models to be able to compare speed-ups for independently-trained vs jointly-trained. 

\textbf{Experiment 3: Real-World 1, Jointly Trained:}
For this Filter-and-Score problem, the full ensemble contains $T=5$ jointly trained base models, and each base model acts on $13$ features out of a total set of $D = 16$
features, where the  feature subsets for the $5$ base models are chosen to maximize the interactions of 
the features in each base model as per \citet{canini:2016}. The train dataset has $183,755$ examples and the test dataset has $45,940$. The task is heavily biased to rejecting candidates: a priori, the probability a sample will be classified negative by the full
ensemble is $0.95$.

\begin{figure}
\includegraphics[width=\linewidth]{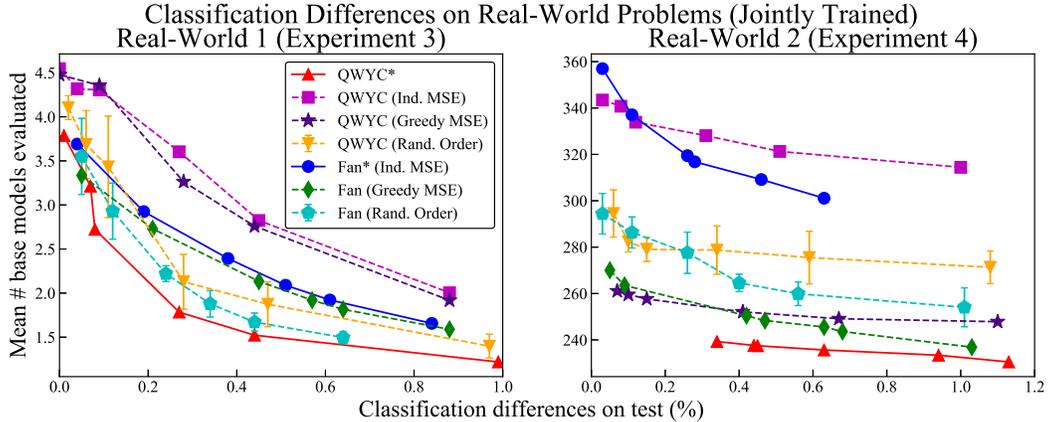}
\caption{Classification differences vs. mean \# base models evaluated for real-world experiments with jointly trained ensembles. The error bars for Random Order show the standard deviation after 5 trials with different random orderings.}
\label{fig:trigger_gbrank_joint_diff}
\end{figure}

\textbf{Experiment 4: Real-World 2, Jointly Trained:}
For this Filter-and-Score problem, the full ensemble contains $T=500$
jointly trained base models, each base model acted on $8$ features out of a total set of $D = 30$ features, and the $500$ feature subsets were randomly generated, which means that some base 
models can be expected to be much more useful than others. The train dataset has $83,817$ examples and the test dataset has $20,955$. The
class priors are roughly equal.

\textbf{Experiments 5 and 6} performed on independently-trained ensembles are detailed in Appendix \ref{sec:ind-real-world}.

An important difference between the real-world datasets and the benchmark datasets is that the the real-world held-out evaluation sets do not always contain the training label, so we do not report accuracies for real-world experiments. This illustrates an important advantage of QWYC*: QWYC* does not require labeled data to optimize for evaluation time, and in production systems, it is often easier to collect unlabeled data that matches the true test distributions than labeled data. 
% In contrast, the method proposed by Fan et al. suggests ordering by total benefits metrics that often require labeled data (such as MSE).

Figure \ref{fig:trigger_gbrank_joint_diff} shows the tradeoff between \% classification differences and mean \# base models evaluated for Experiments 3-6. Since these models are productionized, we also report timing numbers for QWYC* and Fan* at $\approx 0.5\%$ classification differences in Tables \ref{tab:RankingJoint} and \ref{tab:TriggerJoint} in Appendix \ref{sec:more-experiments}. 
This amount of allowed difference was chosen based on a rough estimate of the percent of examples for which the ground truth label is likely noisy.
%The test mean $\mu s$ evaluation time reported is the model's mean evaluation time on a single example across the test dataset.
We report the mean and standard deviation of the per-example evaluation time on the test set after 100 runs. 

\textbf{Discussion:}
In Experiments 1-5, QWYC*'s joint optimization of ordering and early stopping thresholds outperformed any pre-selected ordering combined with Algorithm \ref{alg:2} or \citet{Fan:2002} early stopping mechanisms in terms of tradeoffs between \# base models evaluated and \% classification differences and accuracy (Figures \ref{fig:nomao_adult_acc}, \ref{fig:trigger_gbrank_joint_diff}, \ref{fig:nomao_adult_diff} and \ref{fig:trigger_gbrank_ind_diff} ). In Experiment 6, Random Order outperformed all orderings including QWYC*, and we suspect that this is because for an independently trained ensemble of 500 base models where the base models behave very similarly, a random order is less likely to overfit on a training set than a more ``clever'' ordering like Individual MSE, Greedy MSE, or QWYC*.

Timing experiments on both real world case studies show faster mean evaluation speeds from QWYC* than Fan*, ranging from $1.8$x speed-up for the independently-trained ensemble of $T = 5$ base models in Table \ref{tab:RankingIndependent}, to a notable $4$x speed-up for the jointly-trained ensemble of $T = 5$ base models in Table \ref{tab:TriggerJoint}. 
%Note that since the class priors were approximately equal for Experiment 4 and we would only stop early on negative examples, the amount of speedup is bounded to be not much more than 2x.

In Experiment 1, Fan* performed worse than the baseline of smaller ensembles alone in terms of accuracy and classification differences. We suspect that this is because the Individual MSE ordering did not adequately capture a good cumulative ordering in Experiment 1. We see a similar effect in Experiments 3-6 when the gap between Individual MSE and Greedy MSE is more pronounced with jointly-trained ensembles than independently-trained ensembles.

For many pre-selected orderings, the \citet{Fan:2002} early stopping mechanism performed better than the simple early stopping mechanism in Algorithm \ref{alg:2}. Note that the \citet{Fan:2002} early stopping mechanism is significantly more flexible than Algorithm \ref{alg:2}, with at least 20 thresholds per base model (2 thresholds per bin) rather than just 2 thresholds per base model. This suggests that the QWYC optimization may achieve even better reductions in number of base models if combined with the more flexible early stopping thresholds of \citet{Fan:2002}. However, adding these thresholds increases model complexity and also adds its own evaluation time cost. Even with the simpler early stopping thresholds, QWYC*'s joint optimization of orderings an thresholds was both faster and more accurate than the \citet{Fan:2002} early stopping criteria combined with any pre-selected orderings, which suggests that the joint optimization of ordering and early stopping criteria is valuable.

\section{Conclusions and Open Questions}
We provide the QWYC algorithm, which makes it possible to approximately solve the combinatorial joint optimization of ordering with an early stopping mechanism in practice. QWYC achieves the best approximation bound in polynomial time for certain cases. We have shown experimentally that QWYC's joint optimization of ordering with an early stopping mechanism can produce better results than various pre-selected ordering choices, even when those orderings are combined with nore flexible early stopping criteria.
While we focused here on binary classification, it is straightforward to extend the proposed optimization strategy to multi-class classifiers. Whether a similar strategy can be effectively devised for ranking or regression is an open question. 
While we use a simple early stopping mechanism based on thresholds in our experiments, other pruning mechanisms may be substituted into the QWYC algorithm, and our main contribution is to demonstrate that such joint optimization of the ordering and early stopping mechanism is both valuable and practical.

\label{document:middle}

\newpage
\clearpage
\onecolumn

\appendix

\label{document:appendix}

\showproofstrue

\section{Proofs}\label{app:proofs}

\begin{thm}{pipeline}
For problems in \textsc{Pipeline}, the QWYC algorithm achieves a 4-approximation of the optimal cost, which is also the best achievable approximation bound unless $P = NP$.
\end{thm}

\begin{prf}{pipeline}
% Show that QWYC achieves 4 approx of OPT, and therefore OPT*.
Let $\alpha = 0$ and restrict $S_t(i) = S_t(1)$ regardless of position $i$.
Let $OPT$ be the evaluation time cost of the optimal ordering and thresholds under these restrictions. Finding the optimal $\pi,\epsilon^+,\epsilon^-$ under these restrictions then maps directly onto the Pipelined Set Cover problem, where the goal is to find the optimal ordering $\pi$ such that the ordered sequence of sets $S_{\pi(1)}(1), ..., S_{\pi(T)}(1)$ covers all possible training examples contained in $\cup_{t=1}^T S_t(1)$ while achieving the lowest total evaluation time cost. 

% Let $U = \{e_1,...,e_N\}$ be the set of examples in the training dataset. Let $\{S_1,...,S_T\}$ be the sets of examples that are classified correctly by submodels $\{f_1,...,f_T\}$ respectively. Let the objective be the same as defined in equation \ref{eqn:Xobjective}. When $\alpha = 0$, this objective is equivalent to $\sum_{t=1}^T c_{\pi(t)} |U - \cup_{j=1}^{t-1} S_{\pi(j)}|$. 

Theorem 1 by \cite{Munagala:2005} shows that the Greedy algorithm that chooses the set that minimizes the cost ratio each time achieves a 4-approximation of $OPT$. Since QWYC minimizes the evaluation time ratio $J_r$ when choosing the $r$th base model and early stopping thresholds $\epsilon^+_r and \epsilon^-_r$, QWYC achieves a 4-approximation of $OPT$.

Since in $\textsc{Pipeline}$, $OPT \leq OPT^* + C$, QWYC also achieves a 4-approximation of $OPT^*$ for problems in $\textsc{Pipeline}$. 

\cite{Feige:2004} also show that problems in $\textsc{Pipeline}$ with uniform cost do not admit to better than a 4 approximation unless $P = NP$.
\end{prf}

\subsection{Example of a problem in $\textsc{Pipeline}$}\label{sec:pipeline-ex}
Let $\alpha = 0$ and let $c_t = 1$ for all $t$. Suppose there are examples $\{e_1,...,e_8\}$ and base models $f_1,f_2,f_3$, and the decision threshold is $0$ for the full classifier $\sum_{t=1}^3 f_t$. Let the individual classification decisions for $f_1,f_2,f_3$ be as follows: \\
$f_1(e_1) = 1, f_1(e_2) = -1$ and 0 for all others, \\
$f_2(e_3) = 1, f_2(e_4) = 1, f_2(e_5) = -1$ and 0 for all others, and\\
$f_3(e_5) = -1, f_3(e_6) = 1, f_3(e_7) = -1, f_3(e_8) = -1$ and 0 for all others.\\
For this problem, $S_1(i) = S_1(1) = \{e_1,e_2\}$ regardless of where $f_1$ is evaluated with respect to $f_2$ and $f_3$. The same is true for $S_2(i)$ and $S_3(i)$. Thus, the optimal order is $\pi = [3,2,1]$ and the optimal evaluation time cost is $OPT = OPT^* = \frac{1}{8} (8c_3 + 4c_2 + 2c_1) = \frac{7}{4}$.

\section{Comparison of Different Orderings}\label{sec:order_comparisons}
To illustrate the effectiveness of QWYC*'s joint optimization of ordering and early stopping thresholds in Algorithm \ref{alg:1}, we compare QWYC* to baselines where we combine Algorithm \ref{alg:2} with pre-specified orderings. For completeness of evaluating these orderings, we also compare to hybridized versions of the \citet{Fan:2002} proposal that mixes their early stopping mechanism with other pre-specified orderings. We describe all pre-selected orderings that we use in experiments here:

\textbf{Gradient Boosted Tree (GBT) Ordering:} GBT ensembles are trained in a greedily additive fashion, where each base regression tree $f_t$ is selected to be the one that most improves the training objective when added to the subensemble of size $t-1$. Thus, this training procedure produces a natural ordering of base trees that we compare to.

\textbf{Random Ordering:}
Random ordering of the base models is a random permutation drawn uniformly from all possible permutations of $\{1, \ldots, T\}$. In each experiment, we report the mean and standard deviation of results across 5 random permutations.

\textbf{Order by Individual MSE:}
As done by \citet{Fan:2002}, one can choose the
ordering of the base models $\pi$  based on each base model's individual 
accuracy or MSE, if one has access to labeled examples. For other precision-recall trade-offs, the best individual ordering may be 
slightly different. In preliminary experiments, we found ordering by the models' individual MSE's  to be more useful than ordering by their training accuracies.

\textbf{Order by Greedy MSE:}
Ordering by individual MSE can cause many highly-correlated very-accurate 
classifiers to all be ordered first. Theoretically, a more optimal strategy is 
to first choose the best individual model (by MSE), and then greedily choose 
each subsequent base model to minimize the MSE of the partial ensemble. This
type of greedy ordering is similar to the gradient boosted tree ordering and to other approaches used in non-dynamic
ensemble pruning \citep{MartinezMunoz:2006}.

\section{Details on the Early Stopping Criteria by \citet{Fan:2002}}\label{sec:fan}
We compare to the dynamic ensemble pruning strategy proposed by Fan et al. that they refer to as \emph{dynamic scheduling} \citep{Fan:2002}. Fan et al. recommend that the base models be ordered by decreasing total benefits, where total benefits is suggested to be the mean squared error (MSE) of the individual base model. We refer to this ordering as Individual MSE. 

For a given order of the base models,  Fan et al. choose multiple thresholds per base model, one per bin of incomplete scores \citep{Fan:2002}. They originally apply their method to  averaging ensembles, but it is straightforward to extend their method without modification to additive ensembles. For clarity, we describe our implementation of this in some detail.

Let $g_r(x) = \sum_{t=1}^r f_t(x)$ denote the partial ensemble evaluation up through base model $r$. Let $\beta$ be the decision threshold for classifying the full ensemble evaluation $f(x) = \sum_{t=1}^T f_t(x)$. %Think of $g_r(x)$  as an approximation to the full ensemble evaluation, $f(x) = \sum_{t=1}^T f_t(x)$. Note that because our decision threshold $\beta$ is the decision threshold for the full sum (\ref{eqn:ensemble}), the decision threshold for the full average $F_T(x)$ is $\beta/T$. 

For each base model $r$, define a distinct set $\mathcal{B}_r$ of bins into
which $g_r(x)$ will be mapped, and let $b_r(x) \in \mathcal{B}$ denote the bin that $g_r(x)$ is assigned to. 

For every bin $B_r \in \mathcal{B}_r$, let $\mu_{B_r}$, $\sigma^2_{B_r}$ denote the empirical mean and variance of the
differences compared to the full ensemble evaluation for a set of representative examples that are
mapped to $B_r$. Let $\gamma$ be a confidence hyperparameter that is set by the user to 
control the accuracy of the early stopping with respect to the full ensemble evaluation. Then the early stopping thresholds after the $rth$ base model evaluation and the bin $B_r$ are:
\begin{align*}
\epsilon_{r,B_r}^+ = \mu_{B_r} + \gamma \sigma_{B_r} \: \: \: \textrm{ and } \: \: \: \epsilon_{r,B_r}^- = \mu_{B_r} - \gamma \sigma_{B_r}.
\end{align*}
Fan et al. does not specify a concrete method for how to map $g_r(x)$ to a bin. We found the best and most efficient results using rounding and a knob $\lambda$ to control how many bins there are: bin $g_r(x)$  as  $b_r(x) =
\mathrm{floor}(g_r(x)/\lambda)$, and note $b_r(x)$ is then some $B_r \in \mathcal{B}_r$.  As recommended by Fan et al., we map the bin values to
confidence parameters using a hash table to save evaluation time. For all experiments, we vary the knob $\lambda$ between $0.1$ and $0.001$ to test more or fewer bins, causing the mean number of bins per base model to range from $10$ to $400$. If one uses a large number of bins (small $\lambda)$ we found too few training examples fell into each bin and led to high generalization error, and that a value of  $\lambda = 0.01$ produced the best  tradeoff between speed and accuracy for most experiments.

Thus the decision rule for each example $x$ after the $r$th base model evalution is:
\begin{equation*}
\begin{cases} 
      g_r(x) > \beta + \epsilon_{r,b_r(x)}^+, &   \textrm{classify
      positive} 
      \\
      g_r(x) < \beta + \epsilon_{r,b_r(x)}^-, &   \textrm{classify
      negative} 
      \\
      \textrm{otherwise} &  \textrm{continue evaluating}. 
   \end{cases}
\end{equation*}

It can happen that during evaluation time, an example $x$ is mapped to a bin that does not exist in the hash table if no examples were mapped to that bin on a train set. In that case, \citet{Fan:2002} say to perform the full evaluation for $x$, which is what we do in such cases. In our experiments this happened only for around 10 examples.  

% Maya cut the variant because as you note there are only 10 examples it hardly seems worth the extra worry and I didn't really understand your suggested variant, but suggest we just don't spend time/space on it.
%A variant would be to allow $x$ to be classified early if the bin $b_r(x)$ exists in the hash table for any $r$. 

\section{Additional Experimental Results and Discussion}\label{sec:more-experiments}

\begin{table*}
  \caption{Datasets and Ensembles Used in Experiments}
  \label{tab:datasets}
  \begin{tabular}{lrrrlrl}
 \toprule
Dataset & \# Feat. & Train & Test & Ens. type & Ens. size & Early Stopping \\
\midrule
UCI Adult  & 14  & 32,561 & 16,281 & Grad. boost. trees & 500  & pos. \& neg. \\
UCI Nomao  & 8  & 27,572 & 6,893 & Grad. boost. trees & 500  & pos. \& neg. \\
Real-world 1 & 16 & 183,755 & 45,940 & Lattices & 5  & neg. only \\
Real-world 2 & 30 & 83,817 & 20,955 & Lattices & 500  & neg.  only \\
  \bottomrule
\end{tabular}
\end{table*}

\subsection{Description of Lattice Ensembles}\label{sec:lattice}
For the real-world experiments, we were given ensembles of Lattices. Lattices are similar in evaluation time and flexibility to decision trees, but 
have the advantages that they  produce continuous real-valued outputs $f_t(x) \in
\mathbb{R}$, and result in overall ensemble function $f(x)$ that is smooth and continuous, rather than piecewise-constant \citep{canini:2016}. 
%Also, the models can be constrained to be monotonic with respect to some inputs \citep{canini:2016}, as was the case for the given models. 
The evaluation time for each lattice base model will be roughly equal across an ensemble (since each lattice acts on the same number of features), so we model the base model evaluation costs as $c_t = 1$ for all $t$. 

\subsection{Additional Discussion of Benchmark Experiments 1 and 2}
Figure \ref{fig:nomao_adult_acc} illustrates the tradeoff between the mean number of base trees and the test accuracy for both the Adult and Nomao experiments. Notably, Figure \ref{fig:nomao_adult_acc} includes the baseline where we trained smaller gradient boosted tree ensembles (``GBT alone''). The comparison of QWYC* to GBT alone shows that one can achieve better accuracy by training a larger ensemble of 500 base models and dynamically pruning it down to 40 base models using QWYC* than by training an ensemble of 40 base models in the first place.

\begin{figure}
\includegraphics[width=\linewidth]{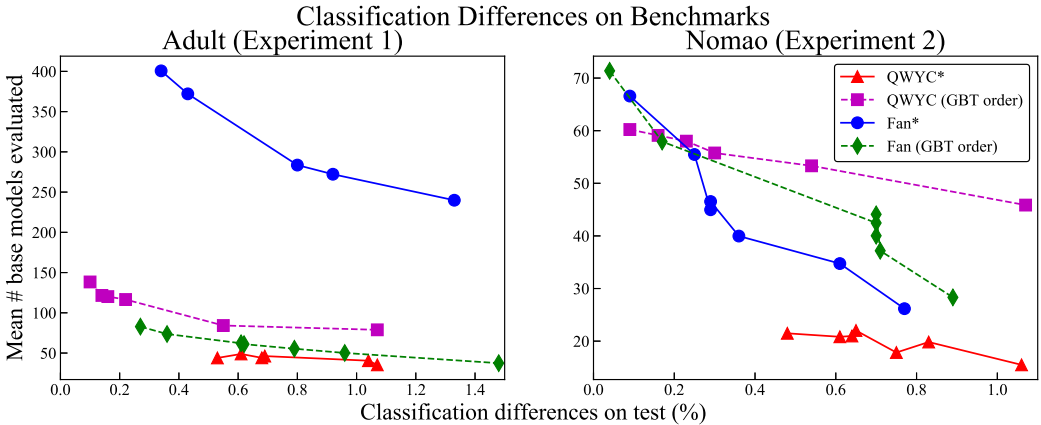}
\caption{Classification differences vs. mean \# base models evaluated for Adult and Nomao experiments.}
\label{fig:nomao_adult_diff}
\end{figure}

\subsection{Independently-trained Real-World Experiments}\label{sec:ind-real-world}
\textbf{Experiment 5: Real-World 1, Independently Trained:}
In this experiment, we use the same setup as Experiment 3, except the base models are independently trained instead of jointly trained. 

\textbf{Experiment 6: Real-World 2, Independently Trained:}
In this experiment, we use the same setup as Experiment 4, except the base models are independently trained instead of jointly trained. 

For both real-world datasets, we see better speedups for
independently trained ensembles than for jointly trained ensembles. This is likely because when independently-trained, each base model correlates more strongly with the full evaluation than when the ensemble is jointly-trained. This effect is most pronounced for real-world dataset 1, with an average $4$x speed-up for the independently-trained ensemble (Table \ref{tab:RankingIndependent}) compared to a $2.7$x speedup for the jointly-trained ensemble (Table \ref{tab:RankingJoint}). The difference is much smaller for real-world dataset 2, but the mean number of base models evaluated is slightly lower for QWYC in Table \ref{tab:TriggerIndependent}.

\begin{figure}
\includegraphics[width=\linewidth]{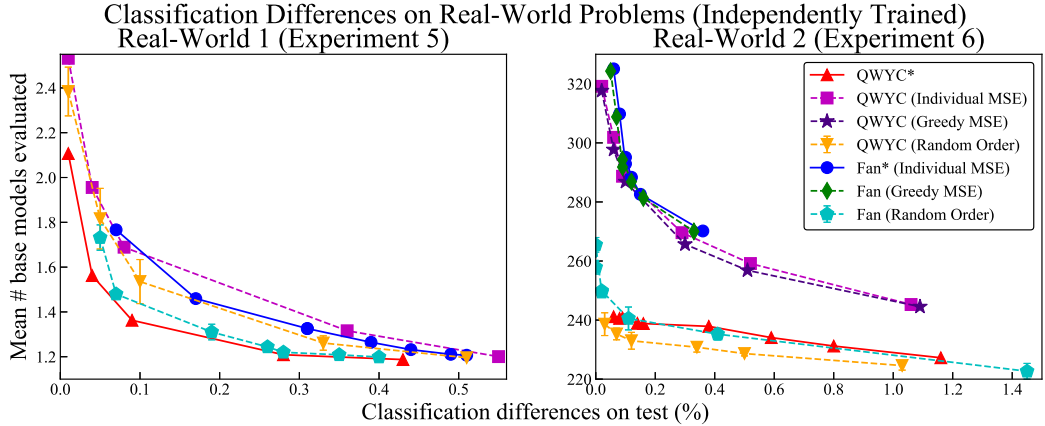}
\caption{Classification differences vs. mean \# base models evaluated for real-world experiments with independently trained ensembles.}
\label{fig:trigger_gbrank_ind_diff}
\end{figure}

For independently trained ensembles, the ordering mattered less, and a random ordering sometimes produced even better speed-ups than the other more "clever" fixed orderings. In Experiment 6, Figure \ref{fig:trigger_gbrank_ind_diff}, the random orderings performed significantly better than both Greedy MSE and Individual MSE orderings. Even in cases where the more "clever" Greedy MSE and Individual MSE orderings failed, QWYC's joint optimization of ordering and thresholds in Algorithm \ref{alg:1} was able to replicate at least statistically similar performance to the best random ordering.

\begin{table}[h!]
  \caption{Experiment 3 -  Comparison of Evaluation Times for Jointly Trained Ensemble of $5$ Base Models.}
  \label{tab:RankingJoint}
  \begin{center}
  \begin{tabular}{lcccccc}
 \toprule
Algorithm & Test   & Test Mean & Test Mean $\mu$s & Test Mean \\ 
& $\%$ Diff& \# Base Models &  Eval Time & Speed-up\\
\midrule
Full ens.  & 0 & 5 & 4.98 $\pm$ 6\% & 1x \\
QWYC           & 0.44\% & 1.54 &  1.82 $\pm$ 7\% & 2.7x \\
Fan & 0.51\%& 2.09 & 2.49 $\pm$ 8\% & 2.0x\\
  \bottomrule
\end{tabular}
\end{center}
\end{table}

\begin{table}[h!]
  \caption{Experiment 4 - Comparison of Evaluation Times for Jointly Trained Ensemble of $500$ Base Models}
  \label{tab:TriggerJoint}
  \begin{center}
  \begin{tabular}{lcccccc}
 \toprule
Algorithm & Test   & Test Mean & Test Mean $\mu$s & Test Mean \\ 
& $\%$ Diff& \# Base Models &  Eval Time & Speed-up\\
\midrule
Full ens. & 0 & 500 & 322 $\pm$ 4\% & 1x \\
QWYC           & 0.45\% & 237 & 177 $\pm$ 4\% & 1.8x \\
Fan & 0.46\%& 309 & 293 $\pm$ 12\% & 1.1x\\
  \bottomrule
\end{tabular}
\end{center}
\end{table}

\begin{table}[h!]
  \caption{Experiment 5 -  Comparison of Evaluation Times for Independently Trained Ensemble of $5$ Base Models}
  \label{tab:RankingIndependent}
    \begin{center}
  \begin{tabular}{lcccccc}
 \toprule
Algorithm  & Test   & Test Mean & Test Mean $\mu$s & Test Mean \\ 
& $\%$ Diff& \# Base Models &  Eval Time & Speed-up\\
\midrule
Full ens. & 0 & 5 & 4.91 $\pm$ 7\% & 1x \\
QWYC           & 0.43\%& 1.18 & 1.23 $\pm$ 6\% & 4.0x\\
Fan & 0.51\%& 1.20 & 1.39 $\pm$ 7\% & 3.5x\\
  \bottomrule
\end{tabular}
\end{center}
\end{table}

\begin{table}[h!]
  \caption{Experiment 6 - Comparison of Evaluation Times for Independently Trained Ensemble of $500$ Base Models}
  \label{tab:TriggerIndependent}
  \begin{center}
  \begin{tabular}{lcccccc}
 \toprule
Algorithm   & Test   & Test Mean & Test Mean $\mu$s & Test Mean \\ 
& $\%$ Diff& \# Base Models &  Eval Time & Speed-up\\
\midrule
Full ens. & 0 & 500 & 269 $\pm$ 4\% & 1x \\
QWYC            & 0.59\% & 234 & 149 $\pm$ 4\% & 1.8x \\
Fan  & 0.46\%& 268 & 209 $\pm$ 7\% & 1.3x\\
  \bottomrule
\end{tabular}
\end{center}
\end{table}

% \subsection{Comparison of Orderings}
% We compared the base-model ordering of QWYC, which is jointly optimized with the early stopping thresholds as described in Algorithm \ref{alg:1},  to  different fixed orders (see Section \ref{sec:order_comparisons}). In all cases, QWYC's optimized ordering produced either the best speed-up or was statistically similar to the best speed-up.

% The second best ordering depended on whether the ensemble
% was jointly-trained or independently-trained. For all of the jointly-trained
% ensembles in our experiments, the Greedy MSE ordering produced the best
% speedups after QWYC (Figures 
% \ref{fig:gbrank_joint_diff} and \ref{fig:trigger_joint_diff}). In a jointly-trained
% ensemble, a single base model is less likely to be representative of the full
% ensemble than in an independently trained ensemble. In fact, the base models in the jointly-trained ensembles had very different outputs, with some base models mostly useful for making positive classifications and others specialized for making negative classifications. The Greedy MSE ordering produced partial ensembles that were clearly better approximations of the full ensemble than the Individual MSE ordering.

\subsection{Distributions of \# base models evaluated per test example}

\begin{figure}
\includegraphics[width=\linewidth]{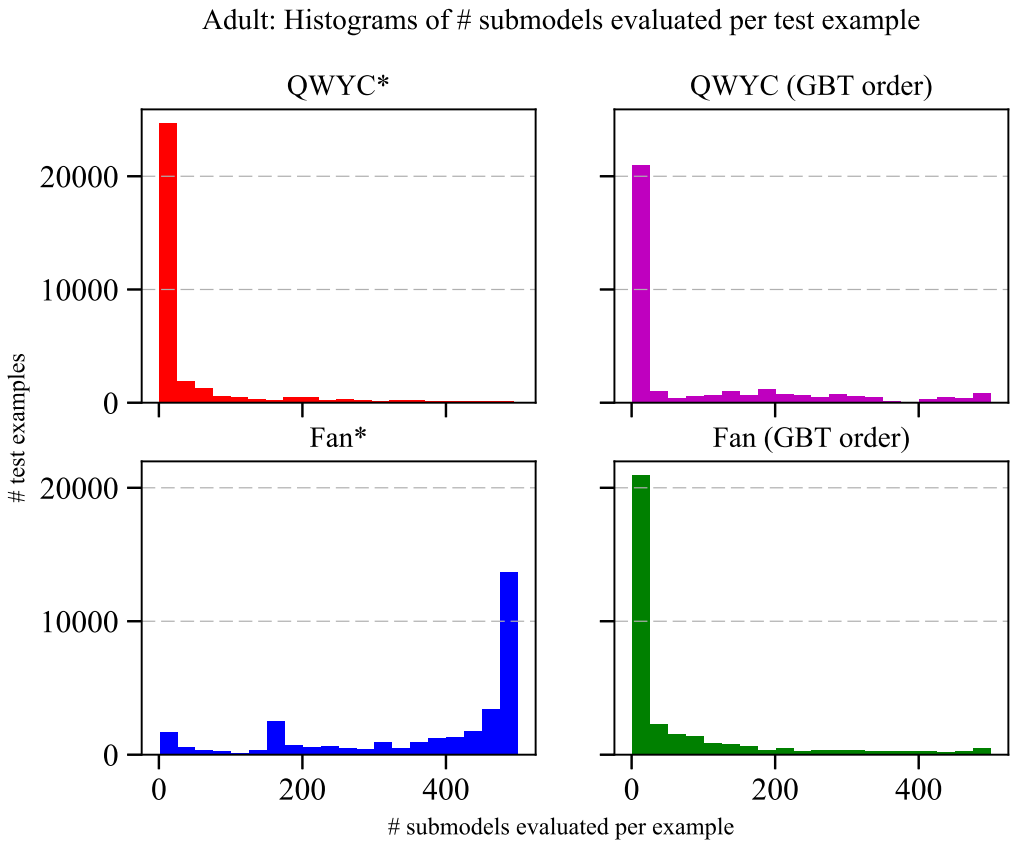}
\caption{Histograms showing the distribution of the number of base models evaluated per example across the test dataset on Adult, for models that exhibit $\approx$0.5\% classification differences to the full ensemble on the test set.}
\label{fig:adult_hist}
\end{figure}

\begin{figure}
\includegraphics[width=\linewidth]{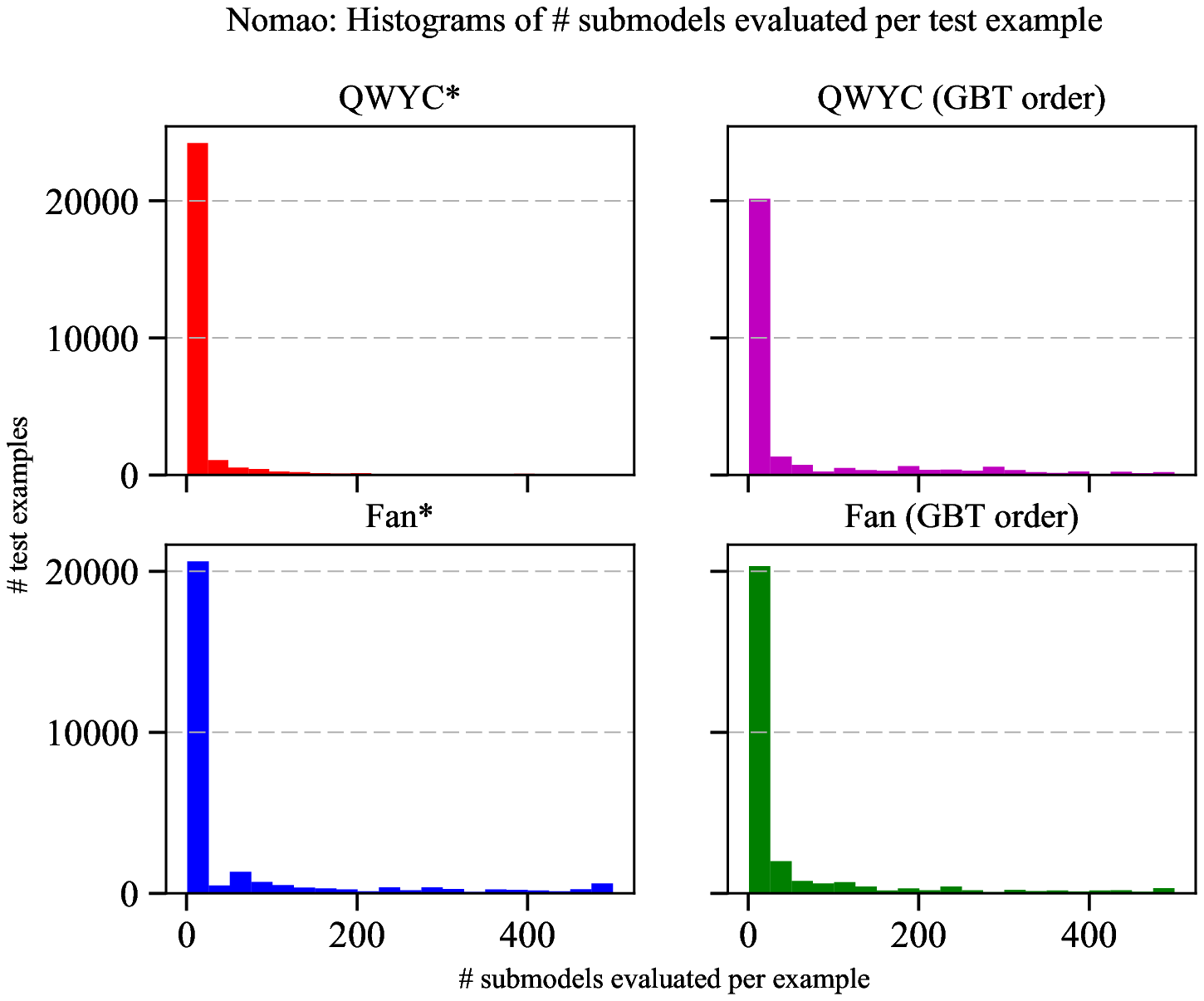}
\caption{Histograms showing the distribution of the number of base models evaluated per example across the test dataset on Nomao, for models that exhibit $\approx$0.5\% classification differences to the full ensemble on the test set.}
\label{fig:nomao_hist}
\end{figure}

Figures \ref{fig:adult_hist} and \ref{fig:nomao_hist} show the distribution across the test set for the number of base models evaluated per example. QWYC exhibits among the smoothest tapering behavior, where the number of examples evaluated with $t$ base models decreases exponentially as $t$ increases.

\label{document:end}

\end{document}